\documentclass[conference]{IEEEtran}
\IEEEoverridecommandlockouts
\usepackage{cite}
\usepackage{amsmath,amssymb,amsfonts}
\usepackage{algorithmic}
\usepackage{graphicx}
\usepackage{textcomp}
\usepackage{xcolor}
\usepackage{booktabs, multirow,array}
\usepackage{float}
\usepackage{subfig}
\usepackage{upgreek}

\def\BibTeX{{\rm B\kern-.05em{\sc i\kern-.025em b}\kern-.08em
    T\kern-.1667em\lower.7ex\hbox{E}\kern-.125emX}}
\begin{document}

\title{Multi-Periodogram Velocity Estimation with Irregular Reference Signals for Robot-Aided ISAC
}

\author{\IEEEauthorblockN{Yi Geng}
	\IEEEauthorblockA{\textit{Cictmobile, China} \\
		gengyi@cictmobile.com}
                            \and
	\IEEEauthorblockN{Pan Cao}
	\IEEEauthorblockA{\textit{University of Hertfordshire, UK} \\
		p.cao@herts.ac.uk}
                	\and
	\IEEEauthorblockN{Ting Zeng}
	\IEEEauthorblockA{\textit{Beihang University, China} \\
		tingzeng@buaa.edu.cn}
        \and
	\IEEEauthorblockN{Yongqian Deng}
	\IEEEauthorblockA{\textit{CMZII, China} \\
		dengyongqian@js.chinamobile.com}
}

\maketitle

\begin{abstract}
This paper addresses velocity estimation within robot-aided integrated sensing and communications (ISAC), where mobile robots act as sensing nodes but can only opportunistically reuse irregular 5G/6G reference signals (RSs). We show that the velocity profile induced by such irregular time-domain patterns can be decomposed into a periodic-peak component and an amplitude-shaping (weighting) component. Leveraging this structure, we propose a multi-periodogram velocity estimation algorithm that is standard-compliant and does not require new sensing-dedicated RSs or 3GPP modifications. Simulation results demonstrate that, compared with conventional periodogram processing, the proposed method improves low-SNR robustness by achieving a 3~dB SNR gain at the 10\% missed-detection rate and reducing false alarms by 51\%.
\end{abstract}


\section{Introduction}
Conventional integrated sensing and communications (ISAC) networks commonly rely on base stations (BSs) and user equipments (UEs) as sensing transmitters/receivers. In monostatic sensing, reliable operation typically depends on maintaining a line-of-sight (LOS) path between the BS and the target~\cite{9585321}; however, in complex environments (e.g., factories or warehouses) with mobile targets, such a BS--target LOS link is often difficult to sustain. In bistatic sensing involving a BS and a UE, a mobile UE can partially alleviate this limitation; nevertheless, sensing performance is often bottlenecked by the UE's limited link budget, restricted array/beamforming gain, and insufficient synchronization and positioning accuracy. In contrast, robot-aided ISAC networks offer a practical means to enhance sensing performance by leveraging mobile connected robots as sensing nodes, since robots can provide stronger RF front-ends as well as improved self-localization and platform stability~\cite{10849725}.

\begin{figure}[t]
	\centering
	\subfloat[]{\label{fig_1a}\includegraphics[width=0.48\columnwidth]{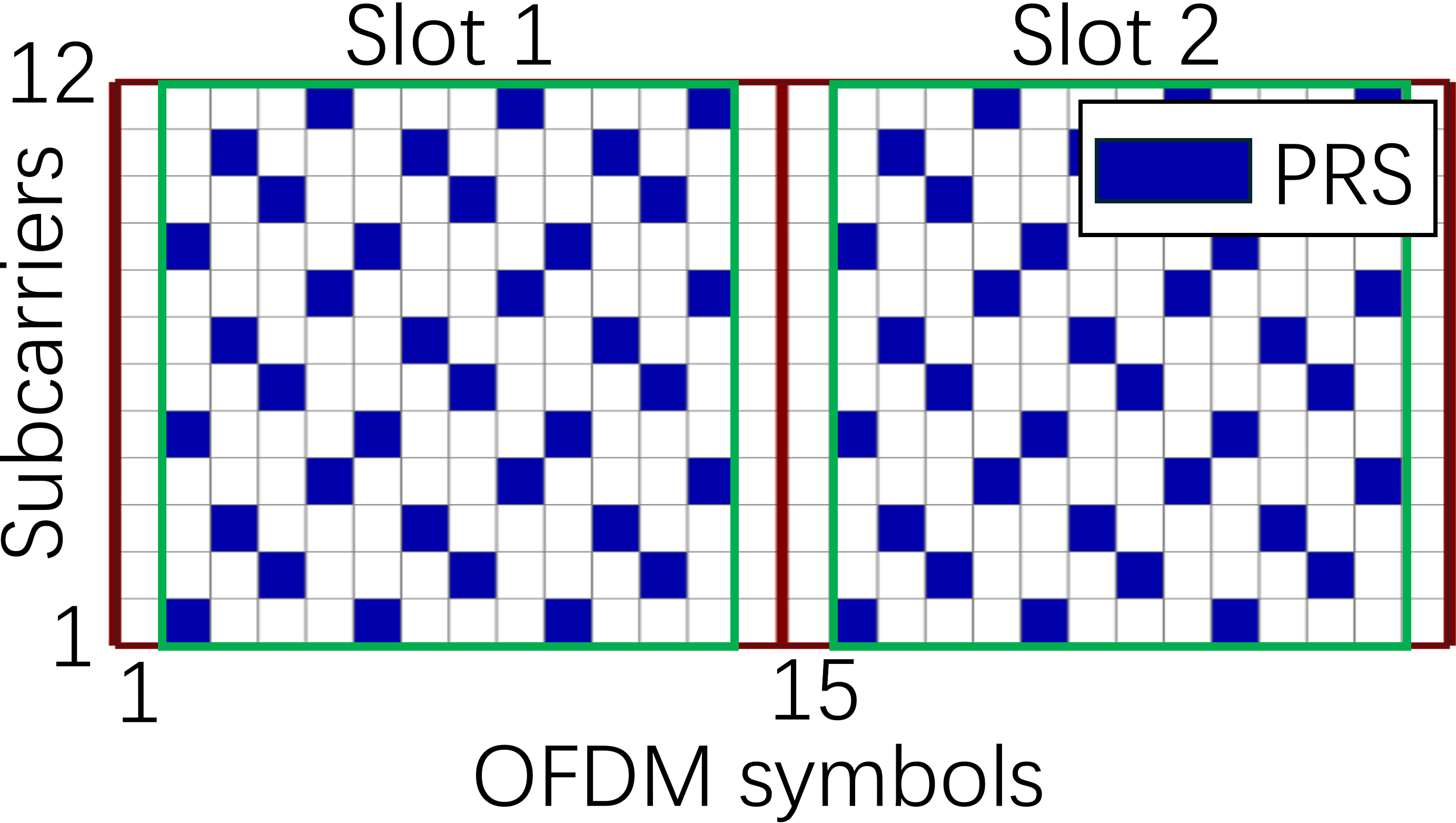}}\quad
        \subfloat[]{\label{fig_1b}\includegraphics[width=0.48\columnwidth]{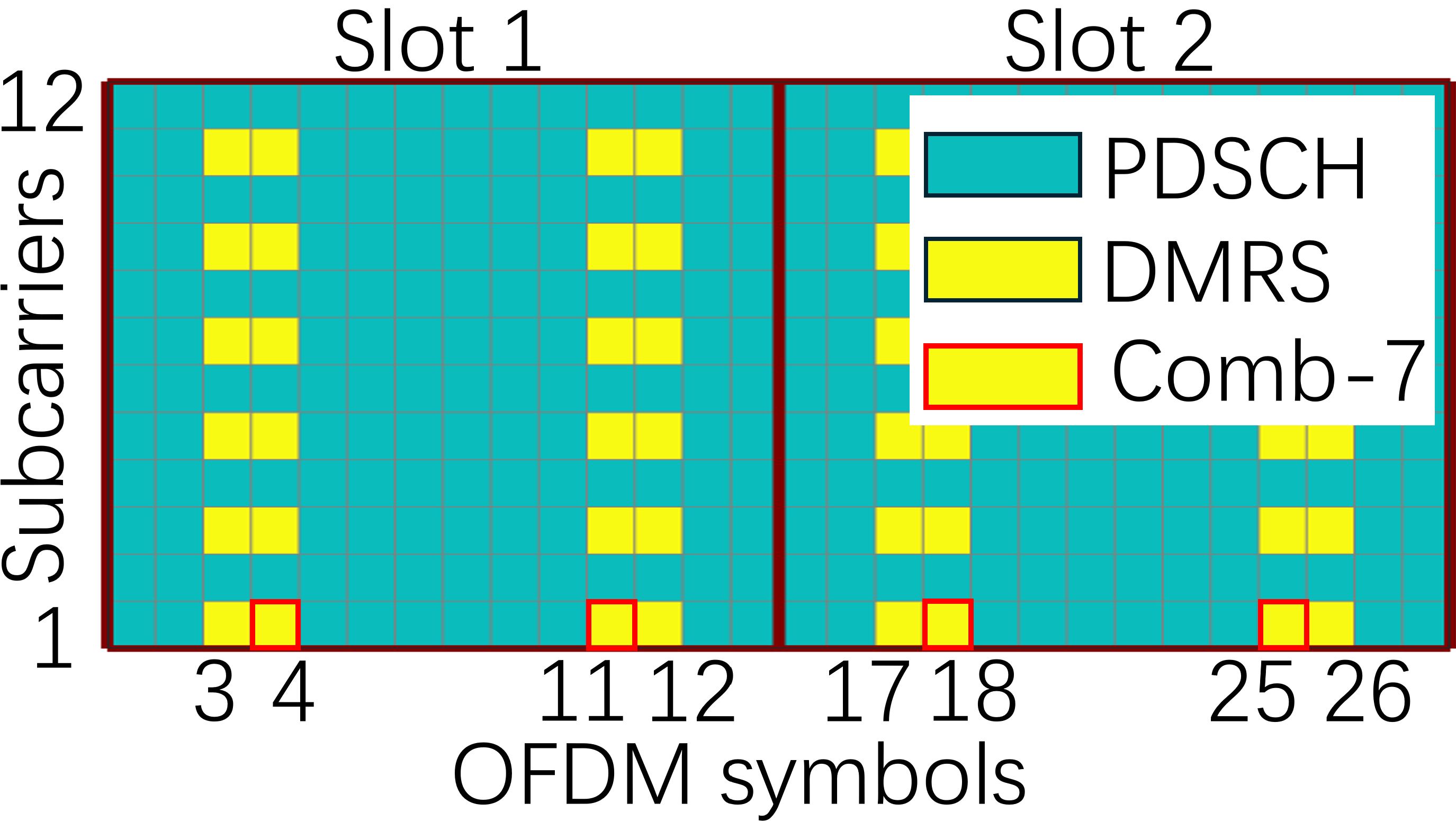}}\quad
        \subfloat[]{\label{fig_1c}\includegraphics[width=0.48\columnwidth]{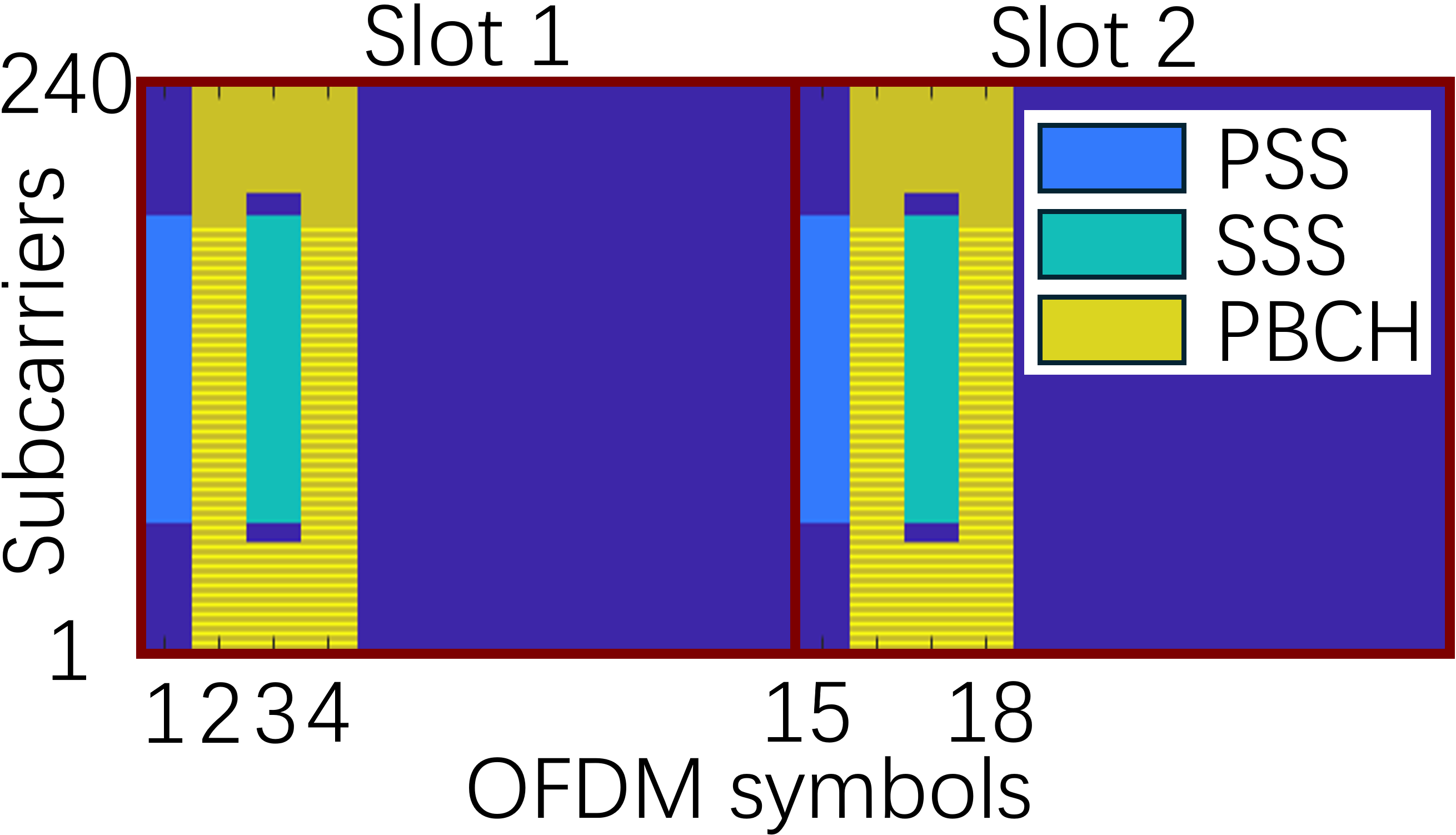}}\quad
        \subfloat[]{\label{fig_1d}\includegraphics[width=0.48\columnwidth]{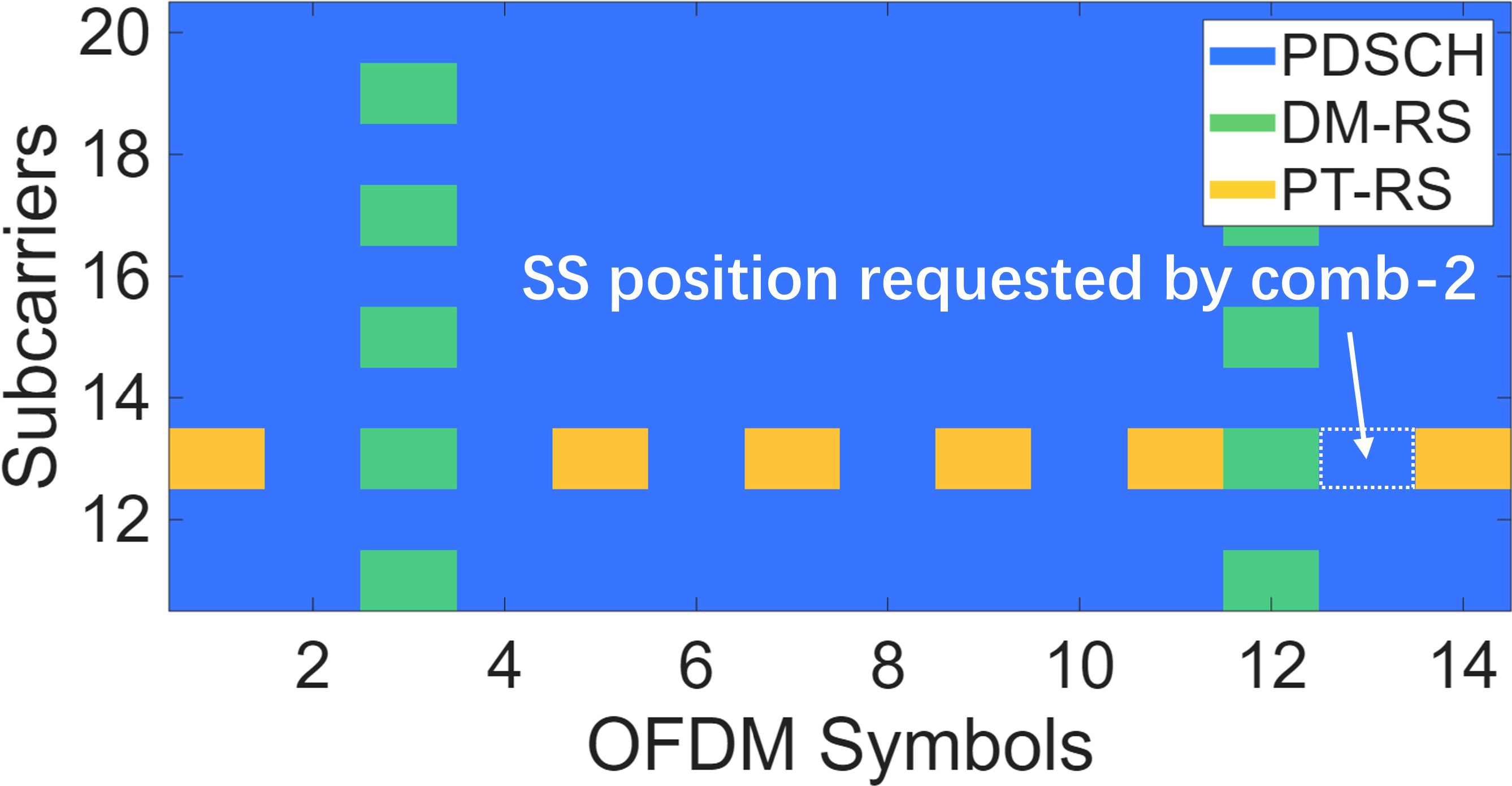}}\\
	\caption{Four SS patterns for velocity sensing: (a) reusing PRS; (b) reusing DMRS; (c) reusing SSB; (d) reusing PTRS.}
	\label{fig_1}
\end{figure}

\begin{figure}[!t]
	\centerline{\includegraphics[width=0.7\linewidth, height=10cm, keepaspectratio]{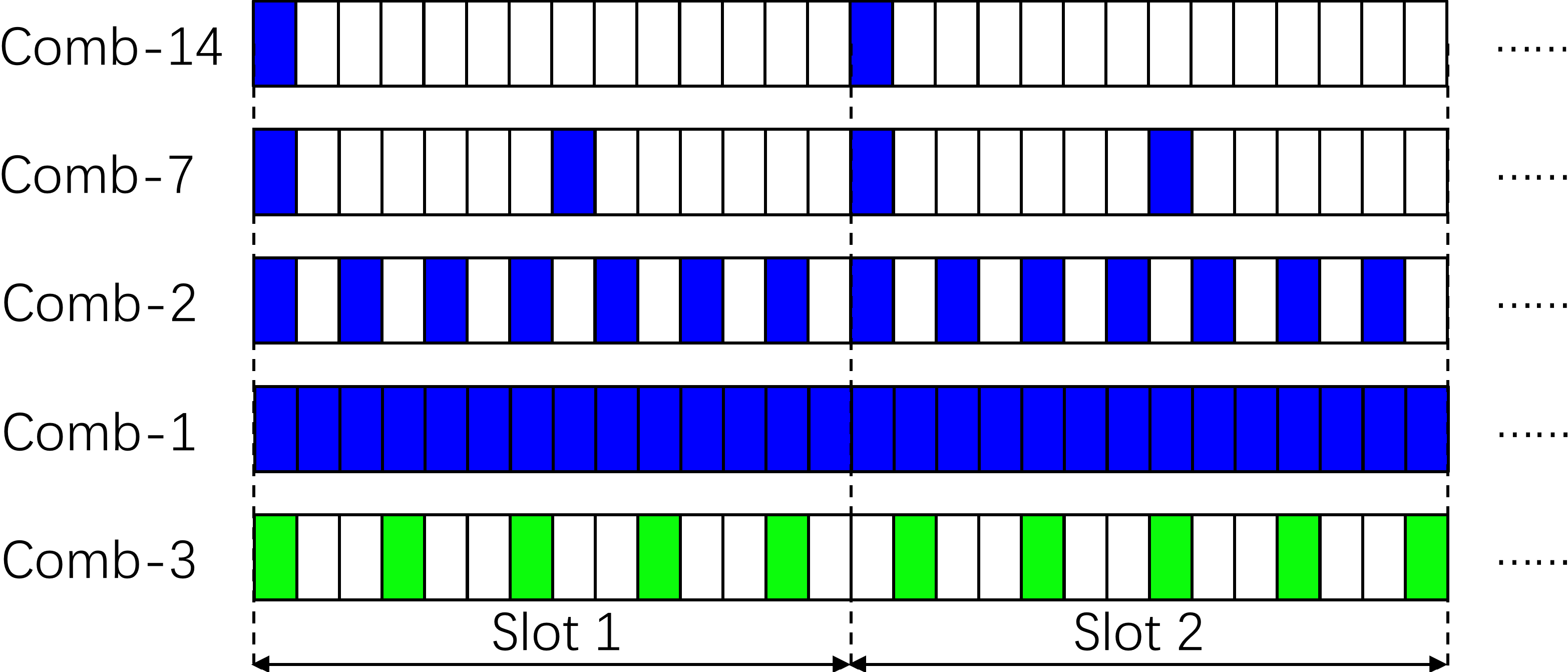}}
	\caption{Comb patterns for velocity estimation.}
	\label{fig_2}
\end{figure}

Recent 3GPP decisions for Release 20 5G-Advanced \mbox{(5G-A)} ISAC indicate that no new reference signals (RSs) dedicated to sensing will be introduced~\cite{3gppRP251798}. This makes the reuse of existing 5G RSs as sensing signals (SSs) a pragmatic and standard-compatible approach~\cite{10788035,10736660,10012421,10833700,10694602}. From a waveform/RS perspective, the structural characteristics of 5G RSs in the frequency and time domains pose different challenges for reuse as SSs. In the frequency domain, many 5G RSs follow regular comb patterns. For instance, the positioning RS (PRS) in Fig.~\ref{fig_1}(a) and demodulation RS (DMRS) in Fig.~\ref{fig_1}(b) exhibit comb sizes of four and two, respectively, which are well-suited for periodogram-based spectral processing that relies on uniformly spaced samples~\cite{9585321}. The key challenge emerges in the time domain, which is crucial for velocity estimation. Since RSs are configured on a per-slot basis and a slot contains 14 symbols, regular time-domain comb patterns can only have comb sizes of 14, 7, 2, or 1 (Fig.~\ref{fig_2}). Table~\ref{tab:table1} summarizes the regular time-domain comb capabilities of typical downlink RSs: synchronization signal block (SSB) is confined to the first 4 symbols of a slot and thus only supports \mbox{comb-14}; PRS occupies 12 out of 14 symbols within each slot (Fig.~\ref{fig_1}(a)), which does allow forming \mbox{comb-7/2/1} over multiple slots; channel state information RS (CSI-RS) is configurable and can support \mbox{comb-14/7/2/1} but is constrained by resource collisions with other RSs; tracking RS (TRS) has a fixed \mbox{4-symbol} spacing within a slot, hence cannot realize \mbox{comb-7/2/1}; DMRS supports \mbox{comb-7} only for specific two-symbol configurations (e.g., Fig.~\ref{fig_1}(b)); and phase tracking RS (PTRS) appears together with DMRS and its \mbox{comb-2} pattern resets at DMRS positions (Fig.~\ref{fig_1}(d)), preventing a true \mbox{comb-2} pattern. As a result, only a limited set of RSs can provide regular comb sampling for velocity sensing.
\begin{table}[!t]
\caption{Regular comb pattern capabilities of 5G RSs\label{tab:table1}}
\centering
\begin{tabular}{|c|c|c|c|c|}
\hline
RS & Comb-14 & Comb-7 & Comb-2 & Comb-1\\
\hline
SSB & $\checkmark$ & $\times$ & $\times$ & $\times$\\
\hline
PRS & $\checkmark$ & $\times$ & $\times$ & $\times$\\
\hline
CSI-RS & $\checkmark$ & $\checkmark$ & $\checkmark$ & $\checkmark$\\
\hline
TRS & $\checkmark$ & $\times$ & $\times$ & $\times$\\
\hline
DMRS & $\checkmark$ & $\checkmark$ & $\times$ & $\times$\\
\hline
PTRS & $\checkmark$ & $\times$ & $\times$ & $\times$\\
\hline
\end{tabular}
\end{table}

This limitation for velocity estimation remains insufficiently addressed in the literature. While~\cite{9540344} recognized the challenge of irregular patterns, it did not provide a concrete solution. The approach in~\cite{9921271} assumes PRS transmission on all symbols (\mbox{comb-1}), contradicting 3GPP standards, whereas~\cite{9449071} uses a very sparse configuration (comb-14) that significantly reduces unambiguous velocity. To fill this gap, we propose a novel \emph{multi-periodogram} velocity estimation framework based on irregular RS patterns, without modifying 3GPP standards for regular SS patterns.

\section{System models}
\subsection{Signal model of regular comb-1 pattern}
To investigate the impact of irregular patterns of 5G RSs on velocity sensing, we start by considering the assumption proposed in~\cite{9921271} that 5G RSs are transmitted across all symbols (comb-1 in Fig.~2) within $N_\text{slot}$ slots. The transmitted signals $d_{\text{TX}}(n)\in \mathbb{C}^{14N_\text{slot}}$ at carrier frequency $f_\text{c}$ are reflected by a target moving with velocity $v$. The echo signals (ESs) $d_{\text{RX}}(n)\in \mathbb{C}^{14N_\text{slot}}$, which experience a stochastic Rician channel model, are received by a connected robot (bistatic sensing mode). The normalized ESs $d(n) \in \mathbb{C}^{14N_\text{slot}}$ are obtained through element-wise division of $d_{\text{RX}}(n)$ by $d_{\text{TX}}(n)$:
\begin{equation}\label{eqn_1}
	d(n) = \frac{d_{\text{RX}}(n)}{d_{\text{TX}}(n)} = \text{e}^{j2\pi \frac{vn}{\Delta v \cdot 14N_\text{slot}}},~n=1,2, \cdots,14N_\text{slot},
\end{equation}
where $\Delta v = \frac{\text{c}_0}{28f_\text{c}T_\text{sym}N_\text{slot}}$ denotes the velocity resolution, $n$ denotes the symbol indices.

Applying fast Fourier transform (FFT) to $d(n)$ yields:
\begin{equation}\label{eqn_2}
	D(k) = \sum_{n=1}^{14N_\text{slot}} d(n)e^{-j2\pi\frac{nk}{14N_\text{slot}}}, \quad k = 1, 2,\cdots, 14N_\text{slot}.
\end{equation}

The velocity profile is obtained by the periodogram of $D(k)$:
\begin{equation}\label{eqn_3}
	\text{Per}(D(k)) = |D(k)|^2, \quad k = 1, 2,\cdots, 14N_\text{slot}.
\end{equation}

A peak appears at bin $k_\text{peak}$ in the velocity profile. The velocity $v$ can be estimated as:
\begin{equation}\label{eqn_4}
	v = \Delta v \cdot k_\text{peak}.
\end{equation}

This approach represents the conventional periodogram algorithm, which is widely employed for velocity estimation in orthogonal frequency division multiplexing (OFDM)-based ISAC systems.

\subsection{Signal model of slot pattern}
As analyzed above, 5G RSs are typically configured on a per-slot basis: while 5G RSs are transmitted on irregular pattern within a slot, the pattern within each slot remains identical across consecutive slots. We define this characteristic pattern as \emph{slot pattern} (\emph{SP}).

Consider an ISAC system where $N_\text{s}$ RSs are transmitted at the $s_1$-th, $s_2$-th, $\cdots$, $s_{N_\text{s}}$-th symbols within each slot, and these RSs are reused as SSs. Fig.~3 illustrates two distinct SPs: SP~1 transmitted at $f_1$ (27~GHz) corresponding to the SSB pattern shown in Fig.~1(c), and SP~2 transmitted at $f_2$ (28~GHz) corresponding to the DMRS pattern shown in Fig.~1(b). The configurations of these two SPs are as follows:
\begin{itemize}
\item{SP~1: $N_\text{s} = 4$ with $s_1=1$, $s_2=2$, $s_3=3$, $s_4=4$},
\item{SP~2: $N_\text{s} = 4$ with $s_1=3$, $s_2=4$, $s_3=11$, $s_4=12$}.
\end{itemize}

$N_\text{slot}N_\text{s}$ RSs are transmitted in $N_\text{slot}$ slots. Let us index these SSs with $m = 1, 2, \cdots, N_\text{slot}N_\text{s}$, where $m = 1,2,\ldots,N_\text{s}$ represent SSs from slot 1, $m = N_\text{s}+1,N_\text{s}+2,\ldots,2N_\text{s}$ represent SSs from slot 2, and so on. For each SS indexed by $m$, we define $l(m)$ as the configuration of the SP:
\begin{equation}\label{eqn_5}
	l(m) = 
	\begin{cases} 
		s_1, & \text{if}~m \bmod N_\text{s} = 0,\\
		~\vdots & \\
        s_{N_\text{s}-1}, & \text{if}~m \bmod N_\text{s} = N_\text{s}-2,\\
		s_{N_\text{s}}, & \text{if}~m \bmod N_\text{s} = N_\text{s}-1.	
	\end{cases}
\end{equation}

\begin{figure}[!t]
	\centerline{\includegraphics[width=1\linewidth, height=10cm, keepaspectratio]{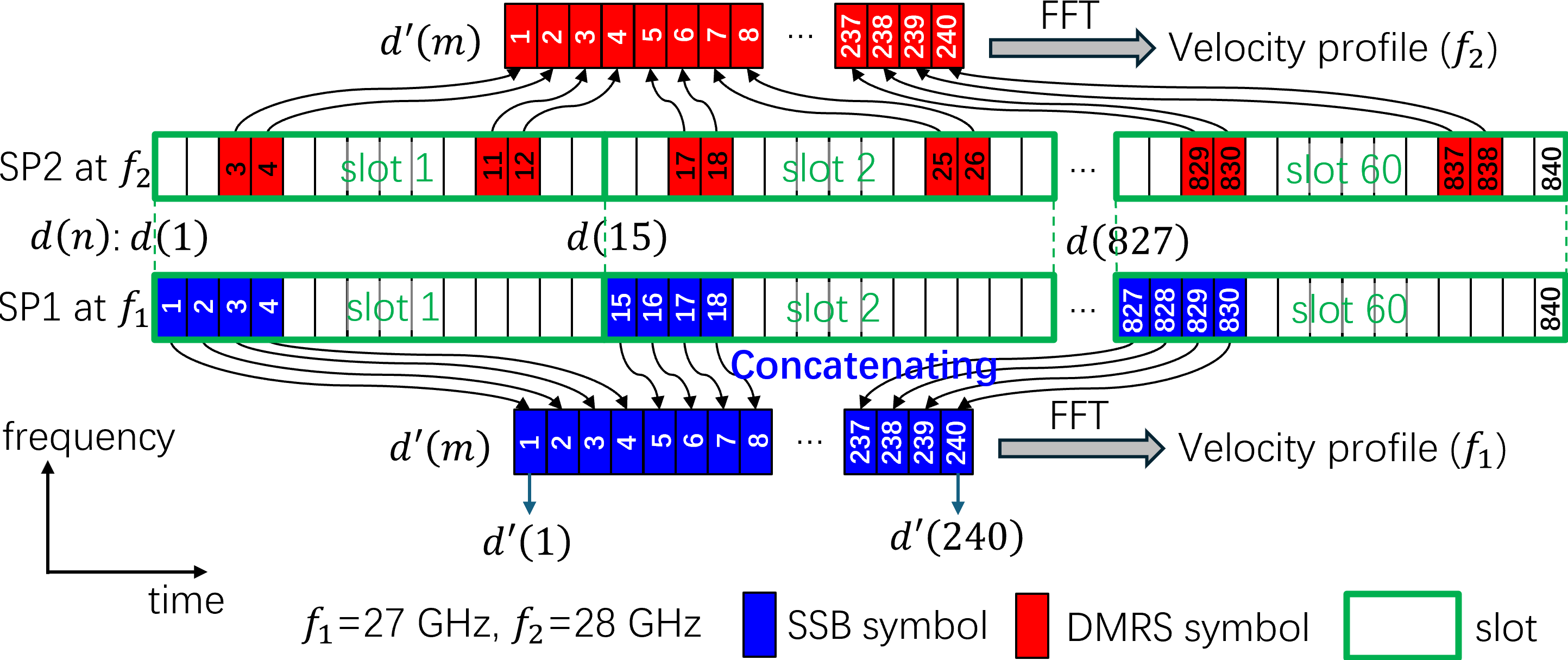}}
	\caption{Illustration of the concatenation process for ESs $d'(m)$.}
	\label{fig_3}
\end{figure}

The SP yields $N_\text{slot}N_\text{s}$ ESs that are discrete in the time domain. A key step in our approach is concatenating these non-uniformly sampled ESs into a vector $d'(m) \in \mathbb{C}^{N_\text{slot}N_\text{s}}$, as illustrated in Fig.~\ref{fig_3}. $d'(m)$ can be expressed as:
\begin{multline}\label{eqn_6}
	d'(m) = d(14\lfloor \frac{m}{N_\text{s}} \rfloor + l(m)) = e^{j2\pi \frac{v}{\Delta v \cdot 14N_\text{slot}}(14\lfloor \frac{m}{N_\text{s}} \rfloor + l(m))},\\m = 1, 2, \cdots, N_\text{slot}N_\text{s}.
\end{multline}

In \eqref{eqn_6}, $\lfloor \frac{m}{N_\text{s}} \rfloor$ ($\left\lfloor \cdot \right\rfloor$ is floor function) identifies the slot index of the \mbox{$m$-th} ES, while $l(m)$ is determined by the SP configuration. The combined term $14\lfloor \frac{m}{N_\text{s}} \rfloor + l(m)$ maps each ES to comb-1's equivalent position in \eqref{eqn_1}. As illustrated in Fig.~\ref{fig_3}, $d'(m)$ represents specific samples of $d(n)$, taken at the positions where RSs are actually transmitted.

\section{Proposed multi-periodogram algorithm}
This section proposes a novel algorithm for SP.
\addtolength{\topmargin}{0.03in}
\subsection{Multi-periodogram algorithm and its velocity profile}
Applying FFT to $\text{d}'(m)$ in \eqref{eqn_6} yields:
\begin{multline}\label{eqn_7}
D'(k) = \sum_{m=1}^{N_\text{slot}N_\text{s}} e^{j2\pi \frac{v}{\Delta v \cdot 14N_\text{slot}}(14\lfloor \frac{m}{N_\text{s}} \rfloor + l(m))}e^{-j2\pi\frac{mk}{N_\text{slot}N_\text{s}}}, \\k = 1, 2,\cdots, N_\text{slot}N_\text{s}.
\end{multline}
where $k$ is the bin index of the velocity profile.

Since the slot index remains the same for all $N_\text{s}$ SSs from the same slot, we define $p$ as the slot index, ranging from 1 to $N_\text{slot}$; $q$ as the symbol index of SSs within a slot, ranging from 1 to $N_\text{s}$. Using $p$ and $q$, \eqref{eqn_7} can be reformulated as
\begin{multline}\label{eqn_8}
D'(k) = \underbrace{\sum_{p=1}^{N_\text{slot}} \overbrace{e^{j2\pi \frac{vp}{\Delta v N_\text{slot}}}}^{\text{Term 1}} \overbrace{e^{-j2\pi\frac{pk}{N_\text{slot}}}}^{\text{Term 2}}}_{D'_1(k)}\underbrace{\sum_{q=1}^{N_\text{s}} e^{j \frac{2\pi v \cdot l(q)}{\Delta v \cdot 14N_\text{slot}}}e^{-j\frac{2\pi qk}{N_\text{slot}N_\text{s}}}}_{D'_2(k)},\\k = 1, 2,\cdots, N_\text{slot}N_\text{s}.
\end{multline}

This reformulation provides a clear understanding of how slot number $N_\text{slot}$ and SP configuration $l(q)$ contribute to the velocity profile. Equation \eqref{eqn_8} denotes that the velocity profile $|D'(k)|^2$ is the Hadamard product (denoted by $\odot$) of $|D'_1(k)|^2$ and $|D'_2(k)|^2$:
\begin{equation}\label{eqn_9}
	|D'(k)|^2 = |D'_1(k)|^2\odot |D'_2(k)|^2.
\end{equation}

Term~1 and Term~2 in $D'_1(k)$ in \eqref{eqn_8} cancel each other and result in unity under the condition $\frac{vp}{\Delta v N_\text{slot}} = \frac{pk}{N_\text{slot}}$, creating a peak at bin $\frac{v}{\Delta v}$ in $|D'_1(k)|^2$. Moreover, Term~2 has a period of $N_\text{slot}$, resulting in a periodic structure with identical-amplitude peaks spaced every $N_\text{slot}$ bins, i.e., periodic peaks appear at bin \mbox{$\frac{v}{\Delta v} \pm N_\text{slot}$}, \mbox{$\frac{v}{\Delta v} \pm 2N_\text{slot}$}, $\cdots$. The bin indices of these peaks can be represented by $k(z)$:
\begin{equation}\label{eqn_10}
	k(z) = \text{round}(\frac{v}{\Delta v}) + zN_\text{slot}, \quad z = 0, \pm 1, \pm 2, \cdots.
\end{equation}

Equation \eqref{eqn_10} denotes that the peak at bin $k(0)$ (e.g., $z = 0$) carries the genuine velocity information, while other peaks introduce ambiguities. $D'_2(k)$ in \eqref{eqn_8} is the superposition of $N_\text{s}$ complex exponentials, each represented as $E(q, k)$ with a unique frequency $f(q)$ and initial phases $\theta(q)$:
\begin{equation}\label{eqn_11}
	D'_2(k) = \sum_{q=1}^{N_{\text{s}}} E(q, k) = \sum_{q=1}^{N_{\text{s}}}e^{-j[2\pi f(q)k-\theta(q)]},
\end{equation}
where
\begin{equation}\label{eqn_12}
	f(q) = \frac{q}{N_\text{slot}N_\text{s}},~\theta(q) = \frac{2\pi v \cdot l(q)}{\Delta v \cdot 14N_\text{slot}},\quad q = 1, 2,\cdots, N_\text{s}.
\end{equation}

The superposition of exponentials in \eqref{eqn_11} results in a $D'_2(k)$ whose amplitude varies across different bins. In certain bins, exponentials align in phase (constructive interference), this leads to a significant high amplitude of $D'_2(k)$; when exponentials have opposing phases (destructive interference), they cancel each other out, resulting in a low amplitude of $D'_2(k)$. This phenomenon is analogous to beamforming, where adjusting the wave phases transmitted by individual antennas enables focusing energy in specific directions.

The Hadamard product in \eqref{eqn_9} performs bin-by-bin (e.g., $k$-by-$k$) amplitude multiplication between $|D'_1(k)|^2$ and $|D'_2(k)|^2$, shaping the velocity profile. A key distinction from conventional periodogram algorithms, which yield a single peak for each target, is that our proposed method generates periodically multiple peaks. Based on this, we name our approach the \textit{multi-periodogram algorithm}.

\section{Simulation results}
\begin{table}[t]
	\caption{Parameters and values in the paper\label{tab_1}}
	\begin{center}
		\begin{tabular}{|c|c|c|}
			\hline
			Parameter & Symbol & Values\\
			\hline
			Frequency of SP1 (SSB) & $f_1$ & 27~GHz\\
			Frequency of SP2 (DMRS) & $f_2$ & 28~GHz\\
			Symbol duration & $T_\text{sym}$ & 8.92~$\mu$s\\
			Number of slots & $N_\text{slot}$ & 60\\
			SSs in a slot & $N_\text{s}$ & 4\\
			Velocity resolution at $f_1$ & $\Delta v_{f_1}$ & 0.741 m/s\\
			Velocity resolution at $f_2$ & $\Delta v_{f_2}$ & 0.715 m/s\\
			\hline
		\end{tabular}
	\end{center}
\end{table}
Simulation results are provided in this section to evaluate the performance of the proposed solution. The parameters used in the simulations are listed in Table~II.
\subsection{Velocity profile analysis}
We consider a robot-aided sensing scenario in a factory environment, where a connected robot and an ISAC BS are sensing a Swerling~1 point target with a fixed RCS and a velocity of 80~m/s. ESs undergo two-way free-space path loss, and receiver noise is modeled as complex white Gaussian noise at a prescribed SNR. In the downlink sensing link, the reflected signal further passes through a stochastic Rician multi-path channel plus additive white Gaussian noise. We conduct two separate simulations with distinct SP configurations: one using SP1 shown in Fig.~3, which corresponds to the SSB pattern in Fig.~1(c), at frequency $f_1$ (27~GHz); and the other using SP2 shown in Fig.~3, which corresponds to the DMRS pattern in Fig.~1(b), at frequency $f_2$ (28~GHz); both SPs span 60 slots. The different frequencies lead to different velocity resolutions for the two SPs: $\Delta v_{f_1}$ = 0.741~m/s and \mbox{$\Delta v_{f_2}$ = 0.715~m/s}. Figs.~4(a) and 4(b) depict the velocity profiles for both SPs. Both velocity profiles exhibit periodic peaks with consistent spacing of 60~bins (because $N_\text{slot}$ = 60) between adjacent peaks. In Fig.~4(a), the peak at bin~108 carries the correct velocity information ($108\Delta v_{f_1}$ = 80.03~m/s), while in Fig.~4(b), the peak at bin~112 contains the velocity information (\mbox{$112\Delta v_{f_2}$ = 80.08 m/s}). These profiles validate the periodic structure predicted by our theoretical analysis in \eqref{eqn_10}.

Fig.~4(c) and Fig.~4(d) illustrate the $|D'_1(k)|^2$ component from \eqref{eqn_9} for both SP configurations. These figures display peaks at identical bin positions as those in their respective velocity profiles, but with uniform amplitudes across all peaks. This uniform amplitude characteristic confirms our theoretical analysis that $|D'_1(k)|^2$ depends solely on the target's velocity $v$ and the number of slots $N_\text{slot}$.

Fig.~4(e) and Fig.~4(f) depict the four constituent exponentials ($N_\text{s} = 4$) that form the $|D'_2(k)|^2$ component for each velocity profile. For SP1, $|D'_2(k)|^2$ comprises four exponentials with frequencies $\frac{1}{N_\text{slot}N_\text{s}}$, $\frac{2}{N_\text{slot}N_\text{s}}$, $\frac{3}{N_\text{slot}N_\text{s}}$, and $\frac{4}{N_\text{slot}N_\text{s}}$, with corresponding initial phases $\frac{2\pi v \cdot 1}{\Delta v_{f_1} \cdot 14N_\text{slot}}$, $\frac{2\pi v \cdot 2}{\Delta v_{f_1} \cdot 14N_\text{slot}}$, $\frac{2\pi v \cdot 3}{\Delta v_{f_1} \cdot 14N_\text{slot}}$, and $\frac{2\pi v \cdot 4}{\Delta v_{f_1} \cdot 14N_\text{slot}}$. For SP2, the four exponentials maintain identical frequencies but exhibit different initial phases: $\frac{2\pi v \cdot 3}{\Delta v_{f_2} \cdot 14N_\text{slot}}$, $\frac{2\pi v \cdot 4}{\Delta v_{f_2} \cdot 14N_\text{slot}}$, $\frac{2\pi v \cdot 11}{\Delta v_{f_2} \cdot 14N_\text{slot}}$, and $\frac{2\pi v \cdot 12}{\Delta v_{f_2} \cdot 14N_\text{slot}}$.

Fig.~4(g) and Fig.~4(h) show $|D'_2(k)|^2$ for the two SP configurations, representing the absolute values of the sums of the four exponentials shown in Fig.~4(e) and Fig.~4(f). Different SPs, with their distinct initial phases of exponentials from $|D'_2(k)|^2$, produce varying amplitude distributions across bin positions. For instance, the four exponentials depicted in Fig.~4(e) simultaneously reach their maximum amplitudes near bin~30, resulting in constructive interference. This phase alignment generates significantly higher amplitude in $|D'_2(k)|^2$ around bin~30, as illustrated in Fig.~4(g). Consequently, the Hadamard product (bin-by-bin ($k$-by-$k$) multiplication) between $|D'_1(k)|^2$ and $|D'_2(k)|^2$ produces the velocity profile in Fig.~4(a), where the peak at bin~48 exhibits greater amplitude than other peaks due to this constructive interference effect.
\begin{figure}[t]
	\centering
	\subfloat[]{\label{fig_4a}\includegraphics[width=0.46\columnwidth]{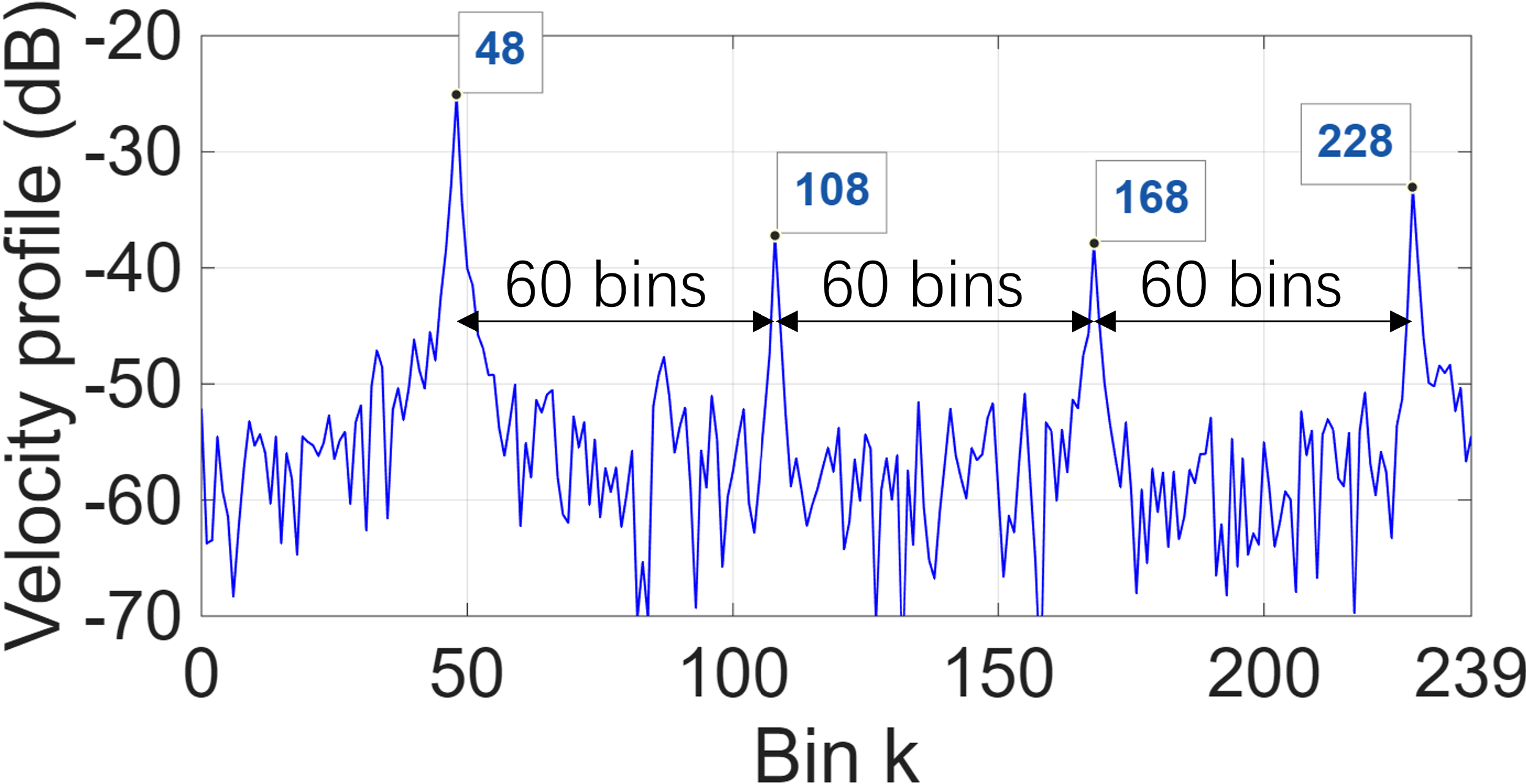}}\qquad
    \subfloat[]{\label{fig_4b}\includegraphics[width=0.46\columnwidth]{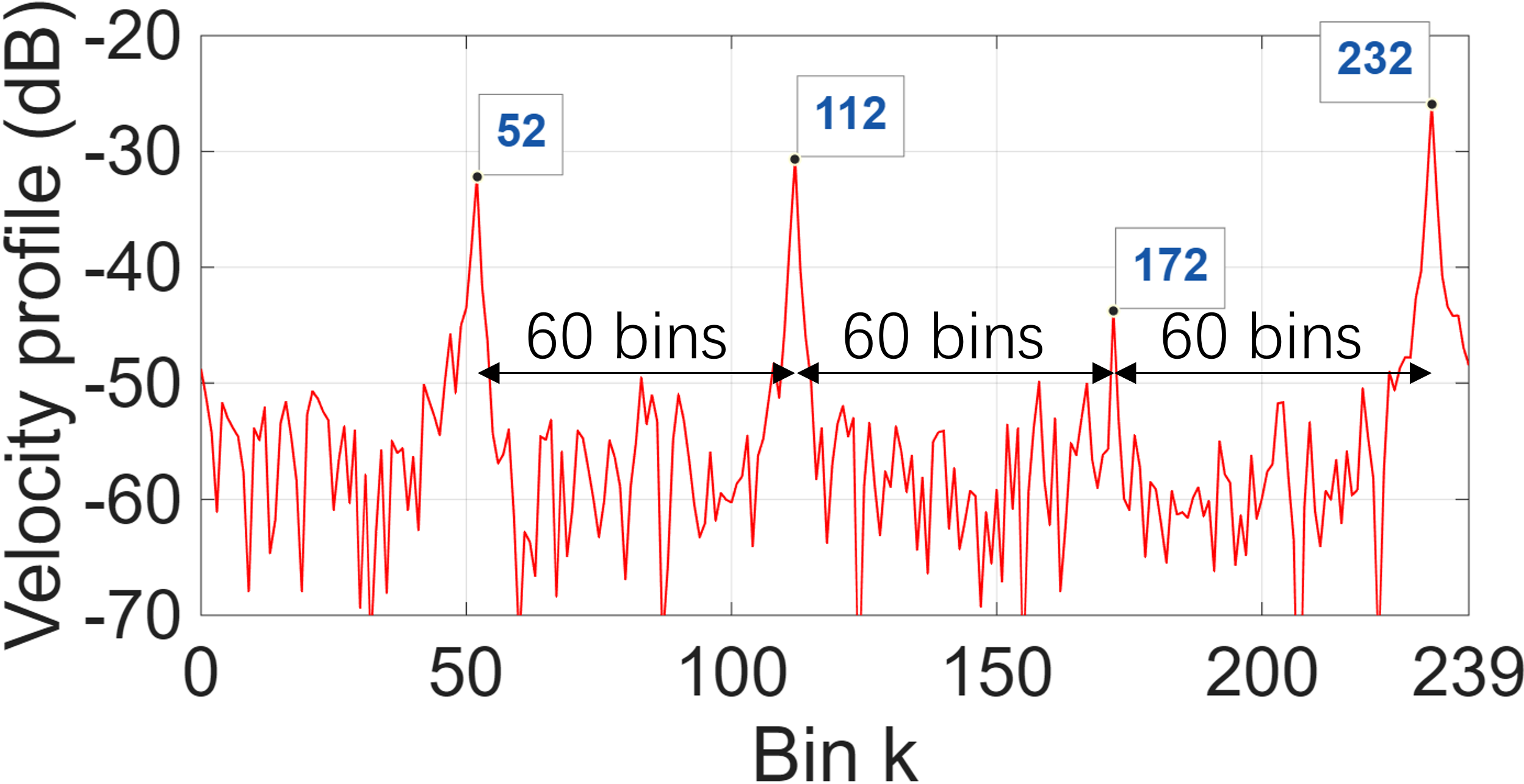}}\\
    \subfloat[]{\label{fig_4c}\includegraphics[width=0.46\columnwidth]{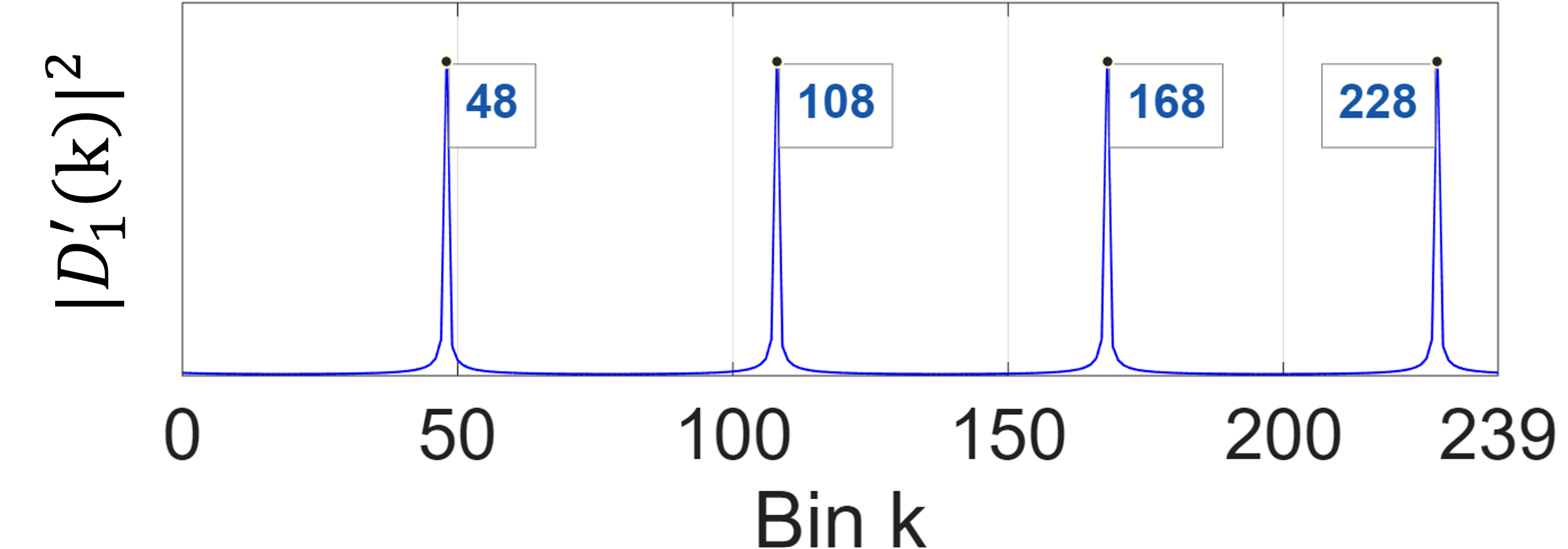}}\qquad
    \subfloat[]{\label{fig_4d}\includegraphics[width=0.46\columnwidth]{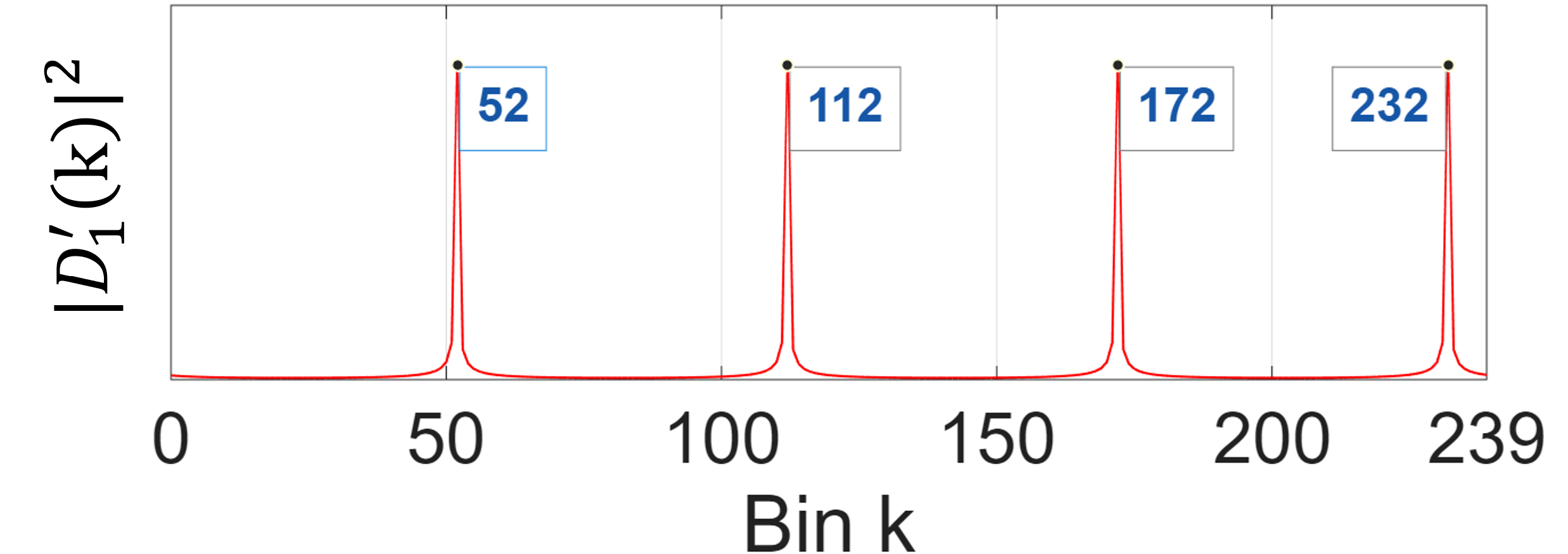}}\\
    \subfloat[]{\label{fig_4e}\includegraphics[width=0.46\columnwidth]{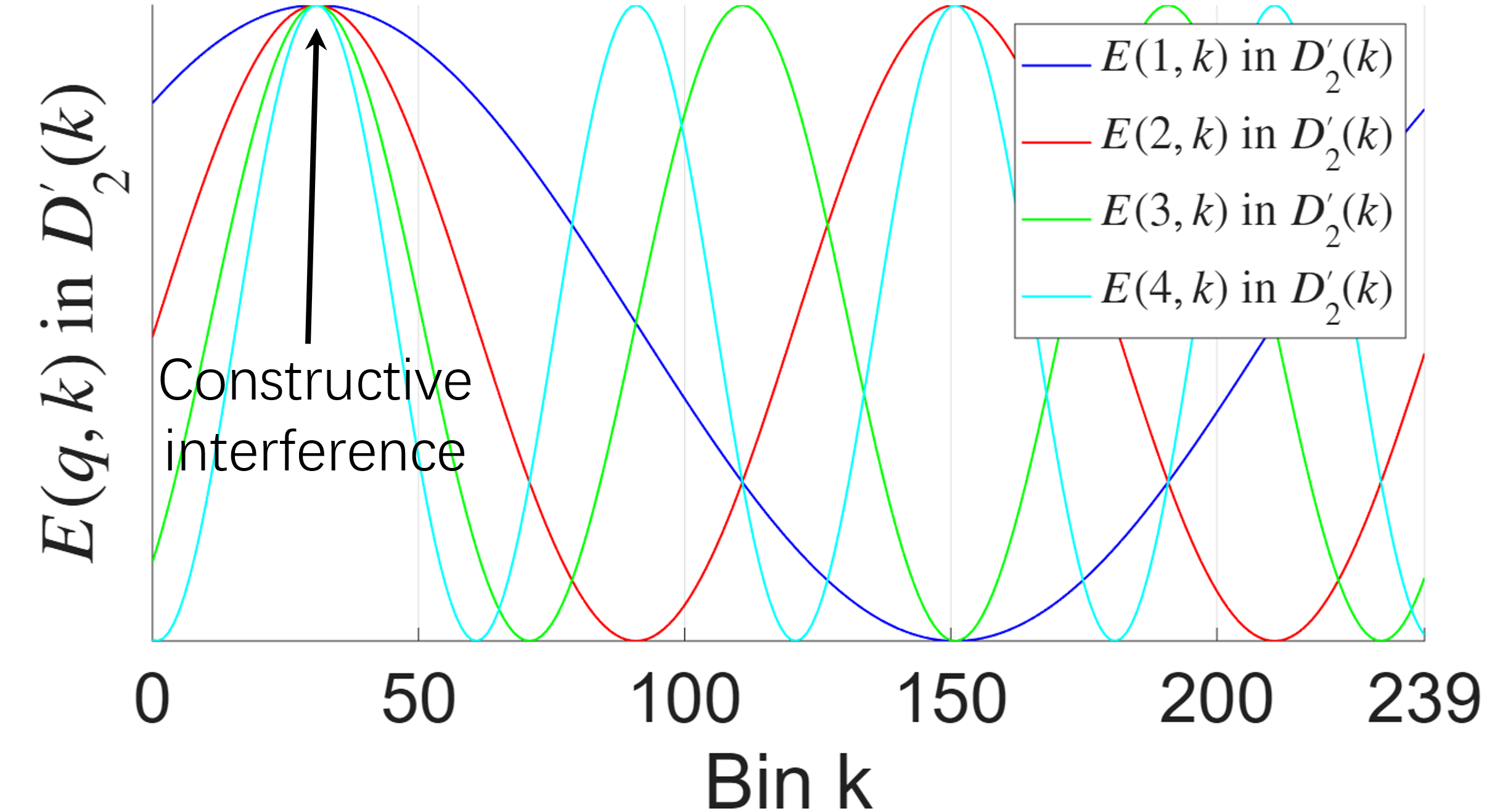}}\qquad
    \subfloat[]{\label{fig_4f}\includegraphics[width=0.46\columnwidth]{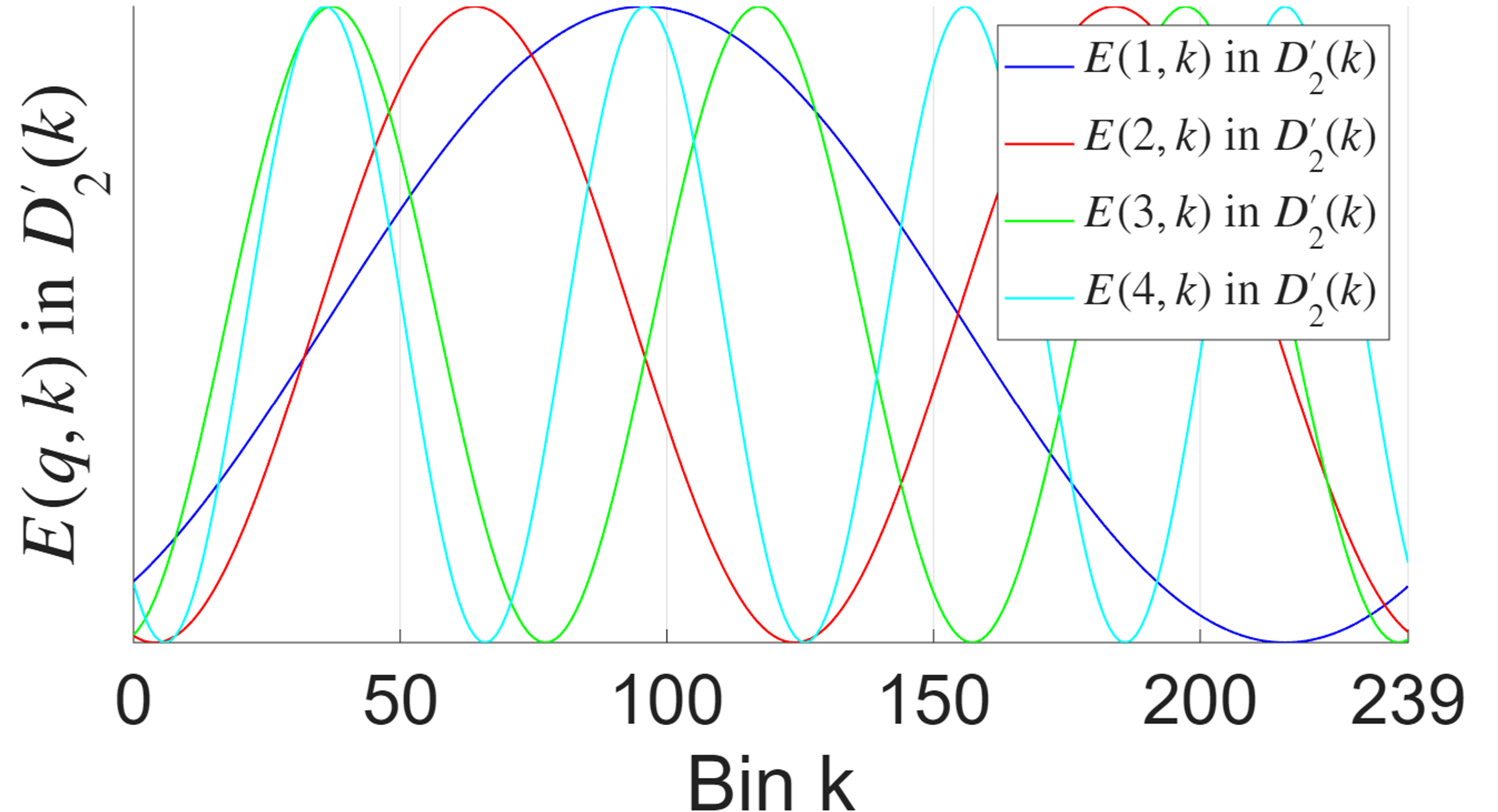}}\\
    \subfloat[]{\label{fig_4g}\includegraphics[width=0.46\columnwidth]{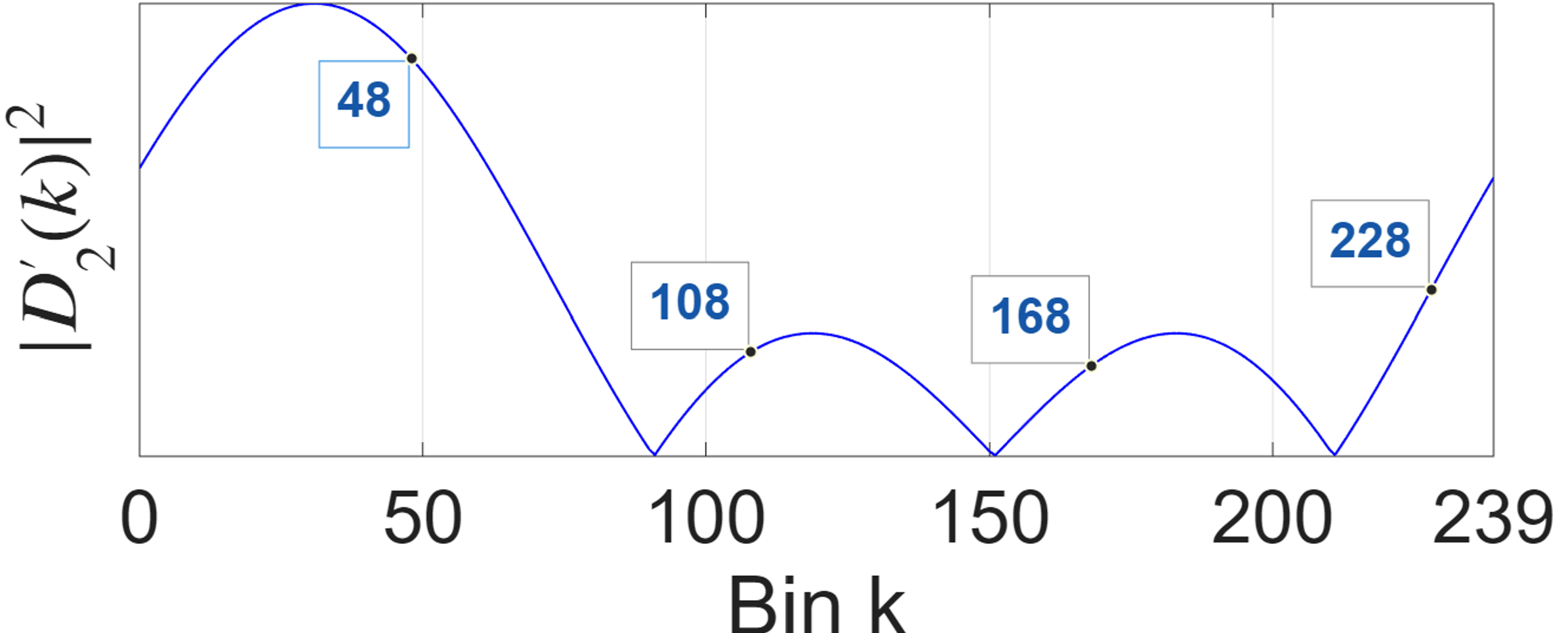}}\qquad
    \subfloat[]{\label{fig_4h}\includegraphics[width=0.46\columnwidth]{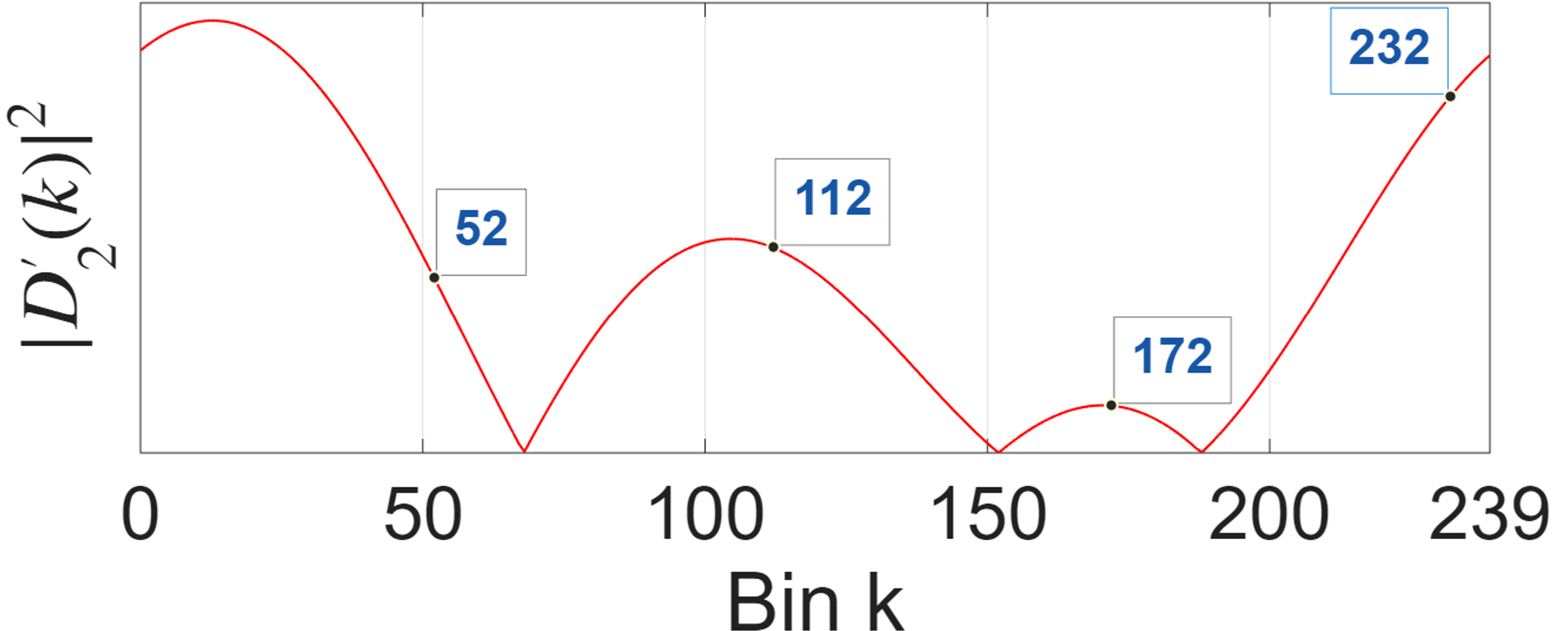}}
	\caption{Analysis of velocity profiles detecting a velocity of 80~m/s. (a) velocity profile derived from SP1 at 27~GHz; (b) velocity profile derived from SP2 at 28~GHz; (c) $|D'_1(k)|^2$ of SP1; (d) $|D'_1(k)|^2$ of SP2; (e) $E(q,k)$ constituting $D'_2(k)$ of SP1; (f) $E(q,k)$ constituting $D'_2(k)$ of SP2; (g) $|D'_2(k)|^2$ of SP1; (h) $|D'_2(k)|^2$ of SP2.}
	\label{fig_4}
\end{figure}

\subsection{Ambiguity mitigation - single target}
As demonstrated above, our proposed algorithm generates periodic peaks in the velocity profile, introducing ambiguity in velocity estimation. To address this challenge, we propose a \textit{multi-frequency peak differentiation method} that mitigates this ambiguity. According to \eqref{eqn_10}, the bin indices of the periodic peaks generated by the target in the two velocity profiles in Fig.~4(a) and Fig.~4(b), denoted as $k_{f_1}(z)$ and $k_{f_2}(z)$, can be expressed as:
\begin{subequations}\label{eqn_13}
\begin{align}
k_{f_1}(z) = \text{round}(\frac{v}{\Delta v_{f_1}}) + zN_\text{slot}, \quad z = 0, \pm 1, \cdots,\label{eq:13A}\\
k_{f_2}(z) = \text{round}(\frac{v}{\Delta v_{f_2}}) + zN_\text{slot}, \quad z = 0, \pm 1, \cdots. \label{eq:13B}
\end{align}
\end{subequations}

For corresponding peaks with identical $z$ values in \eqref{eq:13A} and \eqref{eq:13B}, the bin difference between these peaks in the two velocity profiles remains constant regardless of the value of $z$ since $v$, $\Delta v_{f_1}$, and $\Delta v_{f_2}$ are constants:
\begin{equation}\label{eqn_14}
	\Delta k = k_{f_1}(z) - k_{f_2}(z) = \text{round}(\frac{v}{\Delta v_{f_1}}) - \text{round}(\frac{v}{\Delta v_{f_2}}).
\end{equation}

In Fig.~4(a) and Fig.~4(b), we observe four peak pairs: at bins 48 and 52, bins 108 and 112, bins 168 and 172, and bins 228 and 232, with the two peaks in each pair consistently separated by $\Delta k$ = 4~bins. To identify the peak containing the true velocity information (i.e., the peak where $z = 0$), we select an arbitrary peak pair, e.g., the peaks at bin\mbox{$k_{f_1}(z_0)$ = 48} in Fig.~4(a) and at bin $k_{f_2}(z_0)$ = 52 in Fig.~4(b), where $z_0$ represents an unknown integer. According to \eqref{eqn_10}:
\begin{subequations}\label{eqn_15}
\begin{align}
k_{f_1}(z_0)\Delta v_{f_1} = v + z_0N_\text{slot}\Delta v_{f_1},\label{eq:15A}\\
k_{f_2}(z_0)\Delta v_{f_2} = v + z_0N_\text{slot}\Delta v_{f_2}. \label{eq:15B}
\end{align}
\end{subequations}

From \eqref{eqn_15}, we can solve for $z_0$ = -1 and $v$ = 80~m/s. The value $z_0$ = -1 indicates that the peak pair containing the correct velocity information is located to the first right of our selected pair, specifically at bins~108 and 112. This corresponds to the actual target velocity of 80~m/s, confirming the effectiveness of our multi-frequency peak differentiation method.

\subsection{Ambiguity mitigation - multiple targets}
\begin{figure}[!t]
	\centerline{\includegraphics[width=0.8\linewidth, height=10cm, keepaspectratio]{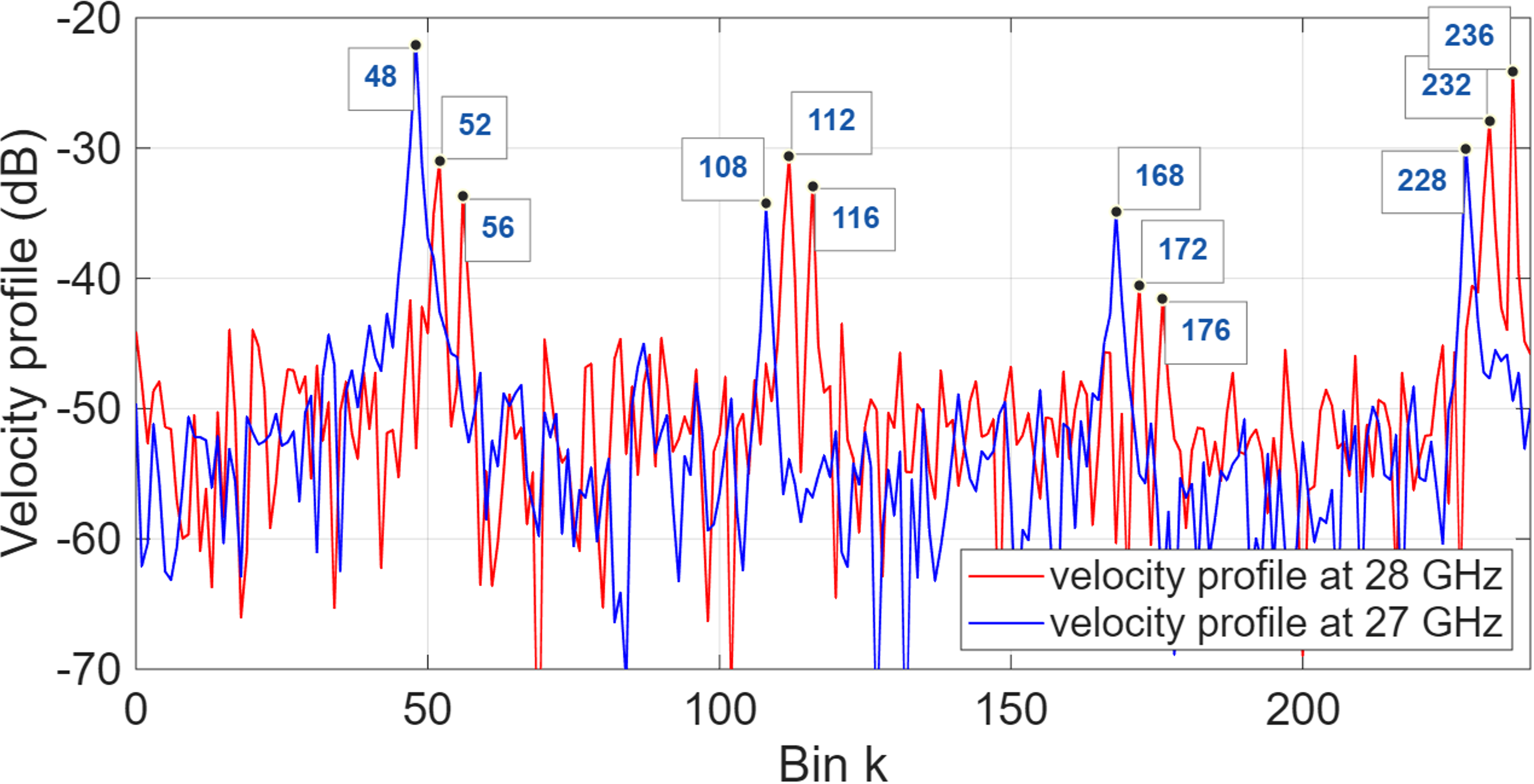}}
	\caption{Velocity profile showing two targets at 80 m/s and 169~m/s.}
	\label{fig_5}
\end{figure}

According to \eqref{eqn_10}, when multiple targets are present with velocities whose values of $\frac{v}{\Delta v}$ differ by integer multiples of $N_\text{slot}$, the proposed algorithm cannot accurately determine the number of targets from a single velocity profile. For example, using the same configuration as in Fig.~4(a) and Fig.~4(b) but sensing two targets with velocities of 80~m/s and 169~m/s respectively, we obtain two velocity profiles, which are plotted together in Fig.~5. The blue curve represents the velocity profile obtained using SP1 at 27~GHz, while the red curve represents the velocity profile obtained using SP2 at 28~GHz.

The blue velocity profile in Fig.~5 appears identical to the velocity profile in Fig.~4(a). This occurs because the peak carrying the 169~m/s velocity information is located at bin~228, which is exactly 120~bins (2$N_\text{slot}$) away from the peak at bin~108 that carries the 80~m/s velocity information. With this single velocity profile alone, it is impossible to determine the number of targets present. However, in the red velocity profile in Fig.~5, the peak carrying the 169~m/s velocity information appears at bin~236, while the peak carrying the 80~m/s velocity information is at bin~112. The distance between these peaks is no longer an integer multiple of $N_\text{slot}$. This distinction enables the disambiguation that is not possible when using a single velocity profile.

\subsection{Superior performance in low SNR scenarios}
\begin{figure*}[t]
	\centering
	\subfloat[]{\label{fig_6a}\includegraphics[width=0.62\columnwidth]{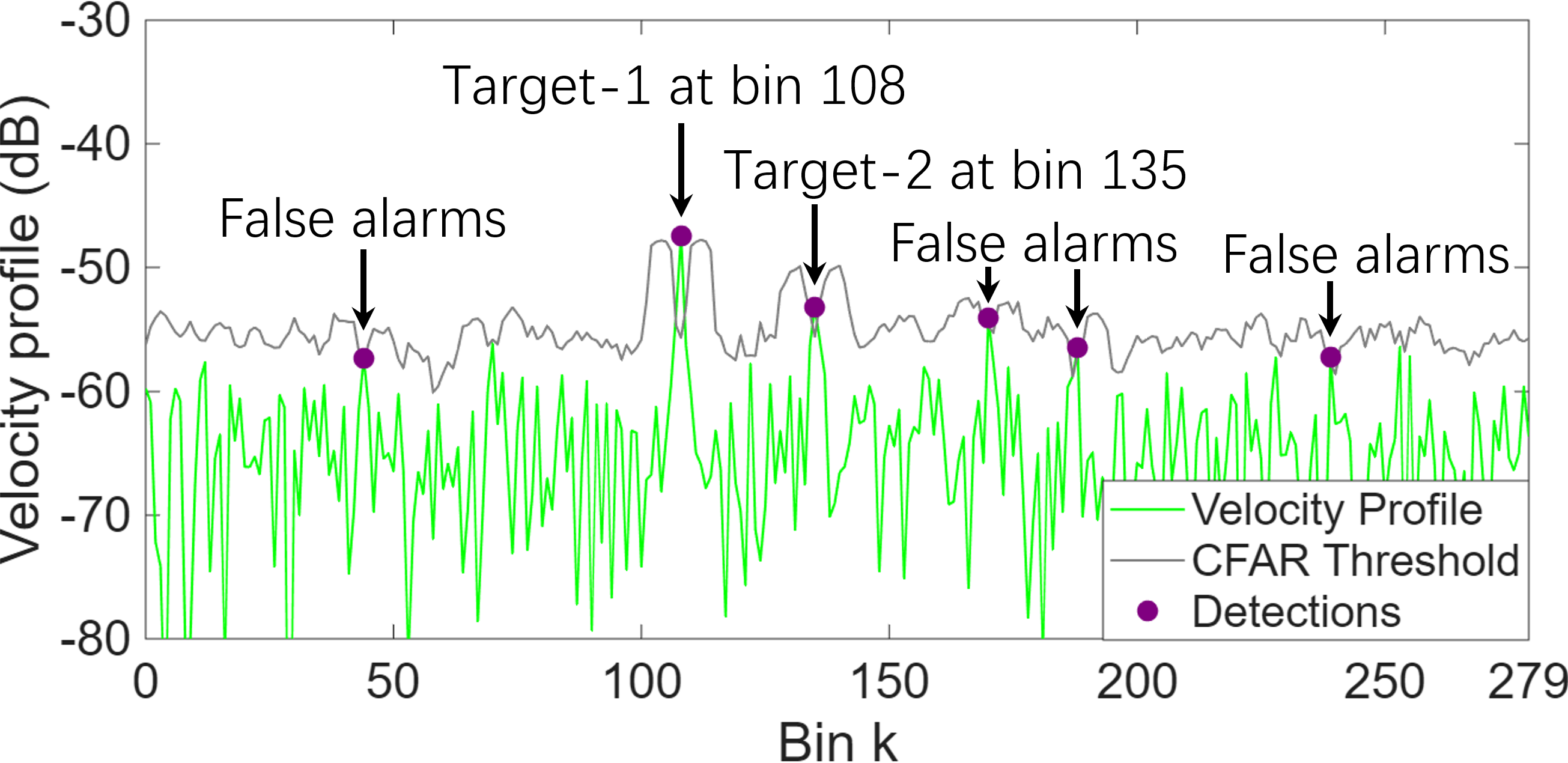}}\qquad
    \subfloat[]{\label{fig_6b}\includegraphics[width=0.62\columnwidth]{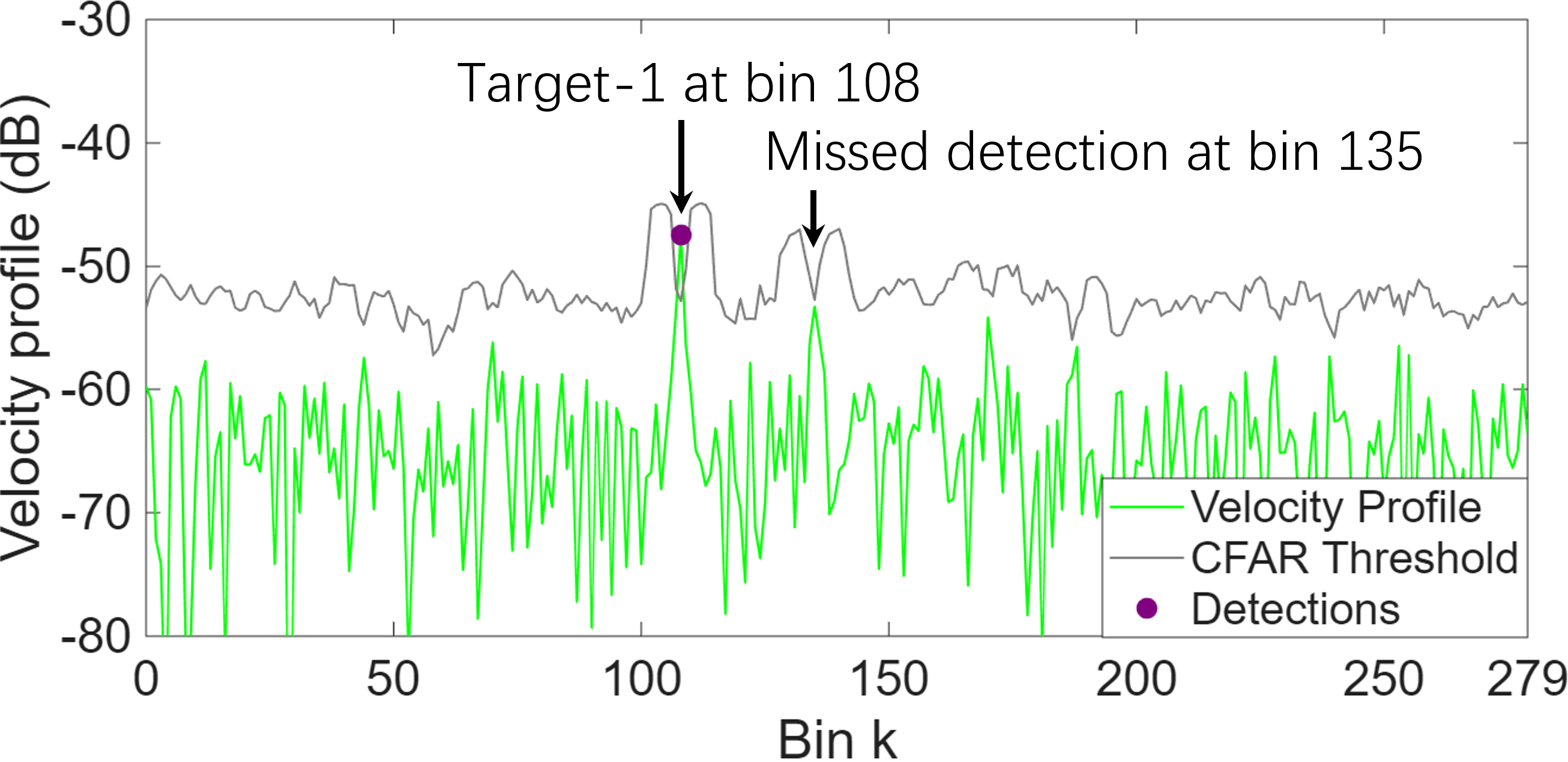}}\qquad
    \subfloat[]{\label{fig_6c}\includegraphics[width=0.62\columnwidth]{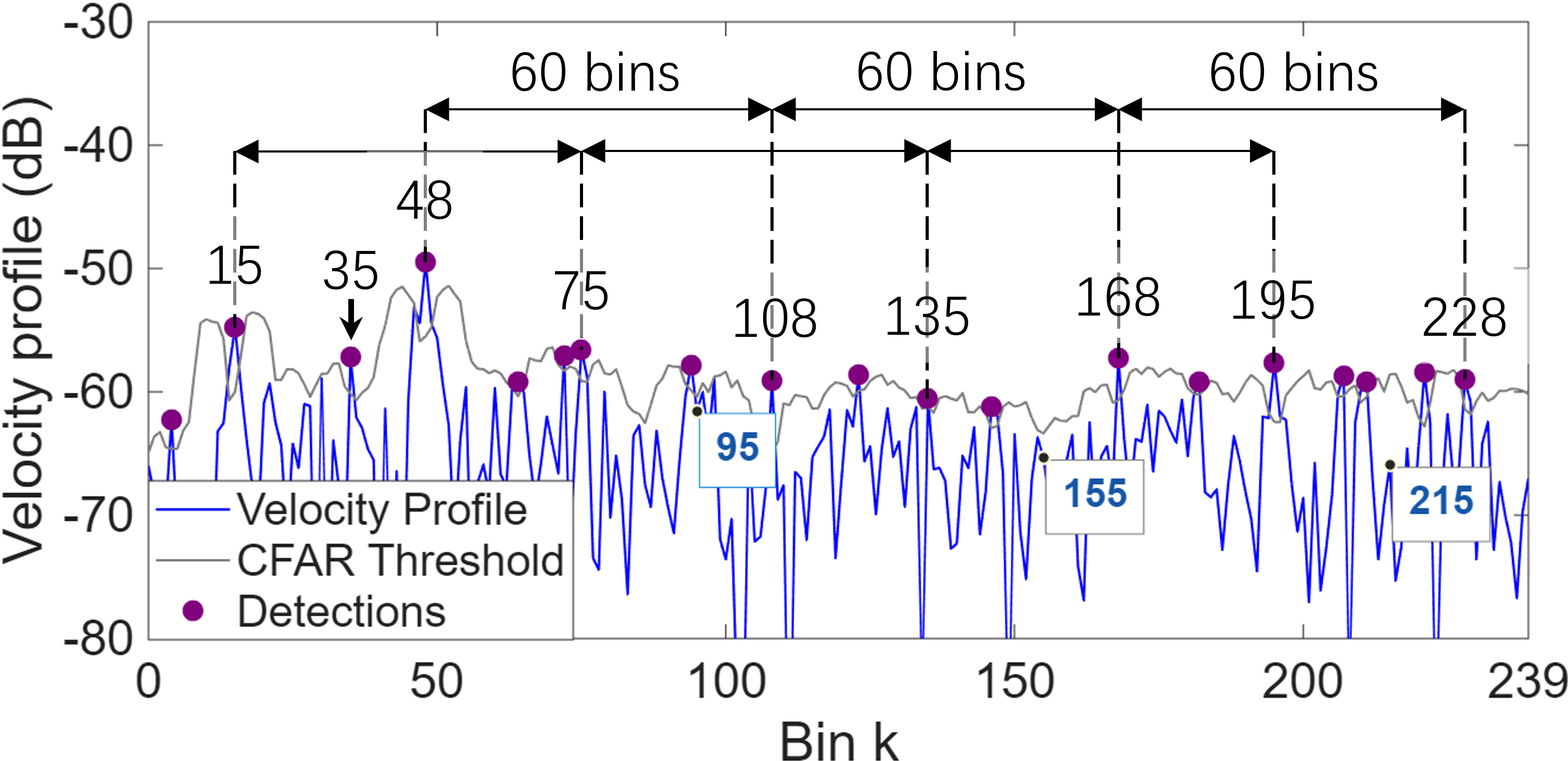}}
	\caption{Detecting two targets with velocities of 80~m/s and 100~m/s. (a) comb-3 pattern and $P_\text{fa}$ of $10^{-2}$; (b) comb-3 pattern and $P_\text{fa}$ of $10^{-3}$; (c) SP1 and $P_\text{fa}$ of $10^{-1}$.}
	\label{fig_6}
\end{figure*}

Compared to conventional periodogram algorithms, our proposed method demonstrates superior performance in low SNR conditions. Using a frequency of 27~GHz and the regular comb-3 SS pattern shown in Fig.~2, we evaluated the detection of two weak targets with velocities of 80~m/s (\mbox{target-1}) and 100~m/s (target-2). Fig.~6(a) and Fig.~6(b) illustrate the velocity profiles with constant false alarm rate (CFAR) detection applied using probability of false alarm ($P_\text{fa}$) values of $10^{-2}$ and $10^{-3}$, respectively, to identify these targets.

Target-2 clearly demonstrates the limitations of periodogram algorithms: to detect this weak target at bin~135, $P_\text{fa}$ must be set to a relatively high value (e.g., $10^{-2}$). However, as shown in Fig.~6(a), this results in numerous false alarms. To reduce these false alarms, when $P_\text{fa}$ is configured to a smaller value (e.g., $10^{-3}$), as illustrated in Fig.~6(b), target-2 can no longer be detected. This occurs because the peak amplitude of weak targets is similar to the noise level, and CFAR methods that rely solely on target peak amplitude cannot simultaneously address both false alarms and missed detections.

In contrast, Fig.~6(c) illustrates the velocity profile obtained using SP1, our proposed algorithm, and a $P_\text{fa}$ of $10^{-1}$. Due to the extremely high $P_\text{fa}$ setting, peaks caused by \mbox{target-1} at bins 48, 108, 168, 228, and those caused by target-2 at bins 15, 75, 135, 195 are all detected. Simultaneously, numerous noise detections with amplitudes similar to the target detections are also identified. The key to distinguishing target detections from noise detections is that peaks caused by actual targets have a fixed spacing of $N_\text{slot}$ bins, while random noise detections rarely satisfy this stringent condition. For example, in Fig.~6(c), if the detection at bin~35 is caused by a real target, detections should also exist at bins 95, 155, and 215. Since no detections exist at these expected positions, we can determine that the detection at bin~35 is caused by noise, and it can be excluded from the false alarm count.

\subsection{Performance evaluation}
\begin{figure*}[t]
	\centering
	\subfloat[]{\label{fig_7a}\includegraphics[width=0.62\columnwidth]{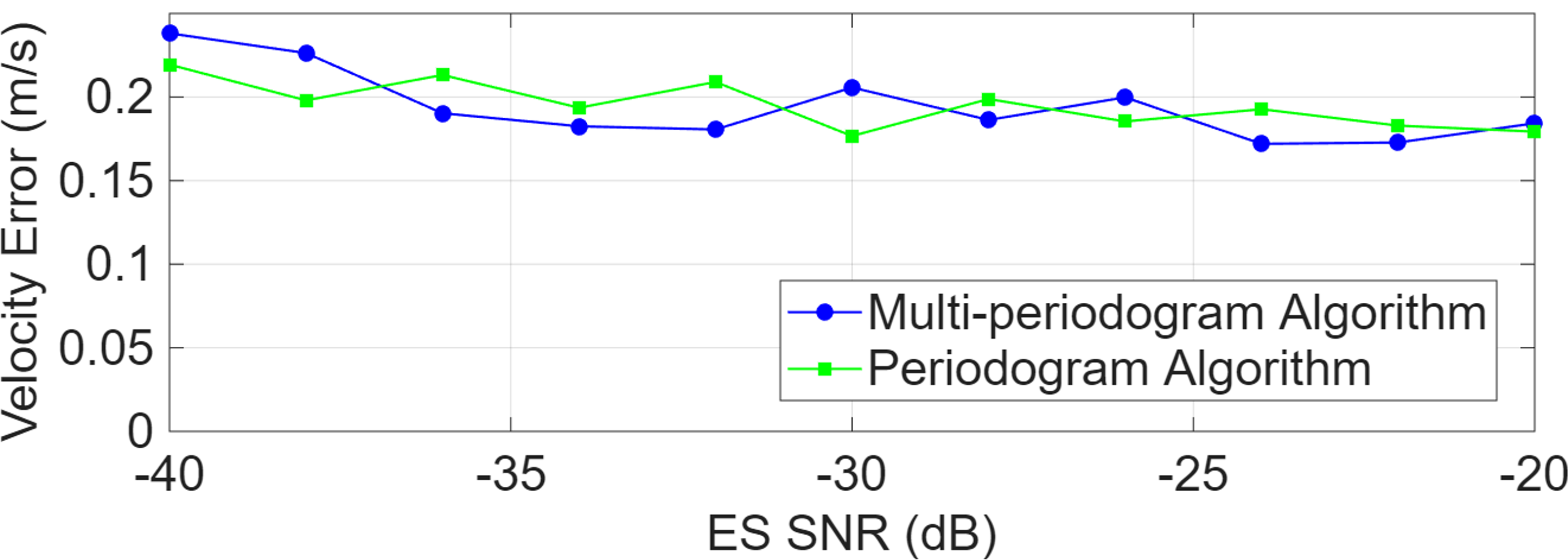}}\qquad
    \subfloat[]{\label{fig_7b}\includegraphics[width=0.62\columnwidth]{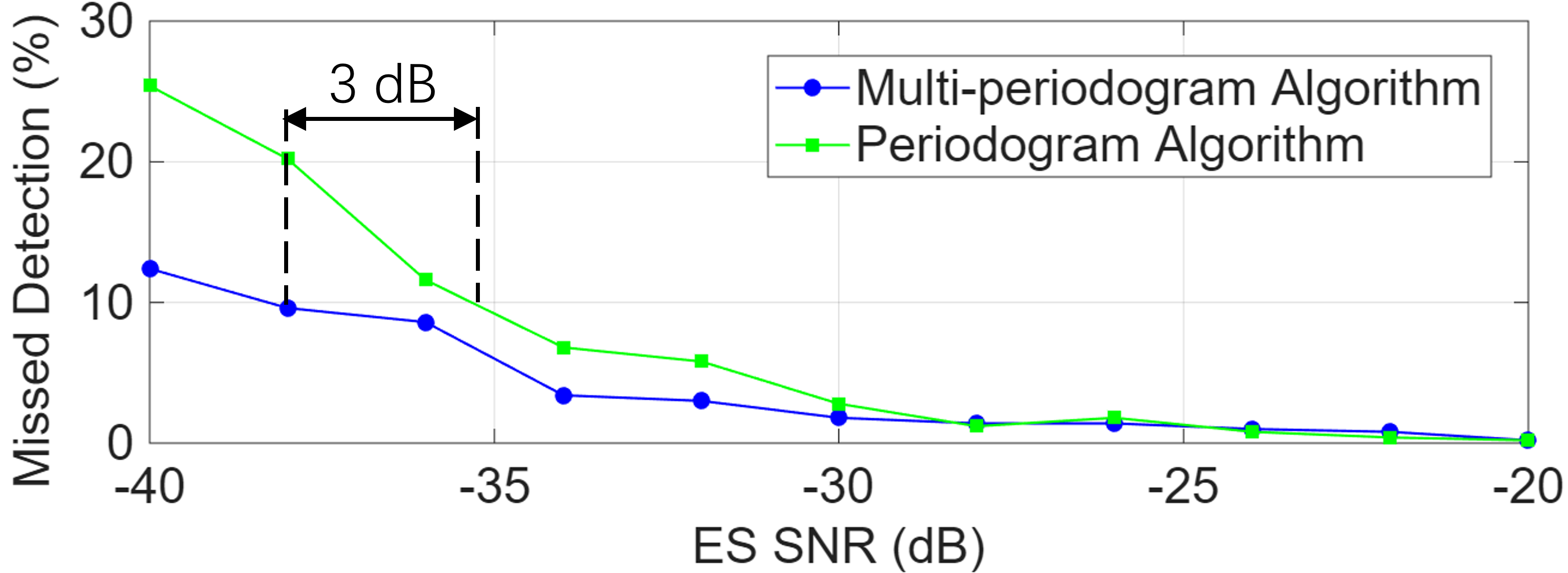}}\qquad
    \subfloat[]{\label{fig_7c}\includegraphics[width=0.62\columnwidth]{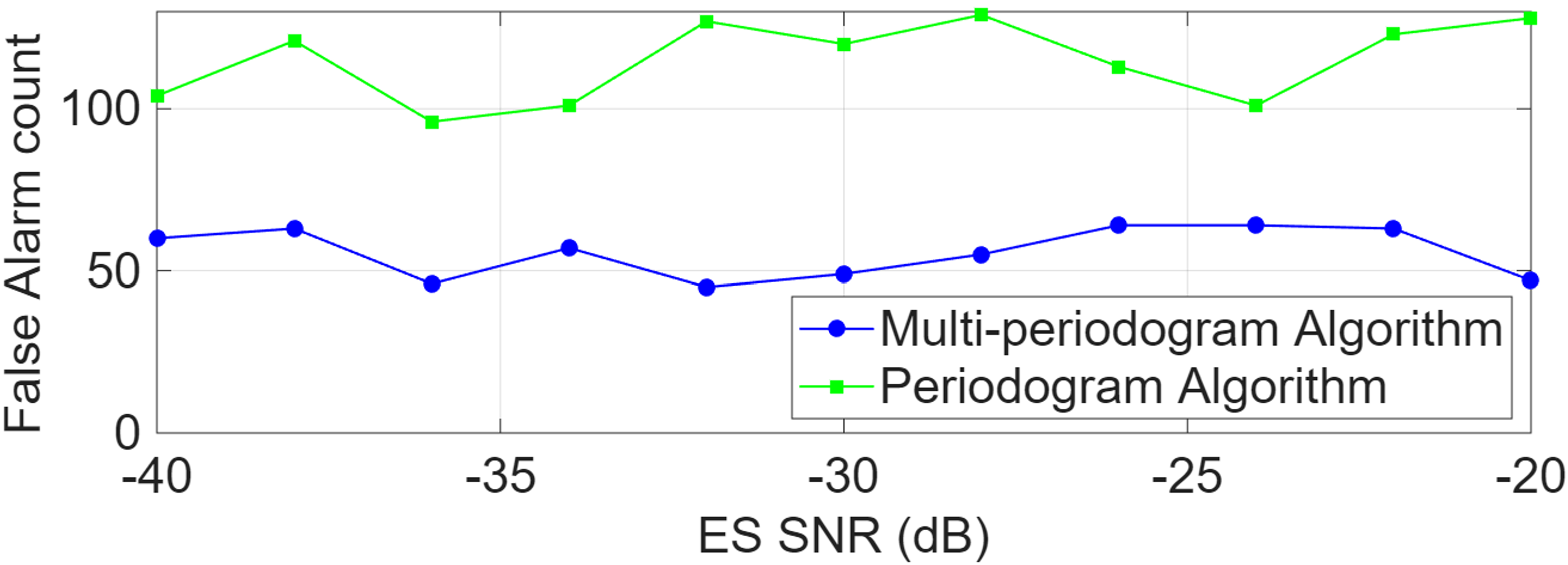}}
	\caption{Performance evaluation between periodogram and multi-periodogram algorithms. (a) velocity error; (b) missed detection rate; (c) false alarm count.}
	\label{fig_7}
\end{figure*}

We conducted Monte Carlo simulations to compare the performance of the conventional periodogram algorithm with our proposed multi-periodogram algorithm. The SSs used by the conventional algorithm followed the comb-3 pattern shown in Fig.~2, while the SPs used by the proposed algorithm are illustrated in Fig.~3. Simulation parameters are listed in Table~II. It's worth noting that the comb-3 pattern used by the conventional algorithm has a higher resource overhead (280 vs. 240 for our approach) and is not supported by 5G-A's RS patterns. Our simulations evaluated velocity error, missed detection rate, and false alarm performance for single-target scenarios under low SNR conditions ranging from -40 to \mbox{-20~dB}. We performed 500 simulation runs for each SNR level, totaling 6,000 simulations.

To balance missed detections and false alarms, the conventional algorithm employed a CFAR parameter with $P_\text{fa}$ of $10^{-3}$, while our proposed algorithm utilized a more aggressive $P_\text{fa}$ of $10^{-1}$. Both algorithms implemented global nearest neighbor (GNN) matching with a 1-bin detection window. Additionally, our proposed algorithm leverages the pattern of periodic peaks to reduce missed detections and false alarms: a detection (whether a potential target or a false alarm) is considered valid only when at least three detections are present at the four expected peak positions in each velocity profile.

Fig.~7(a) shows that both algorithms demonstrate similar velocity error performance. This similarity occurs because GNN matching calculates the velocity error based on the detection closest to the true velocity within the 1-bin detection window, where this velocity error is primarily determined by the velocity resolution, which depends on the signal duration. The performance difference becomes evident in missed detection and false alarm metrics: Fig.~7(b) demonstrates that our proposed algorithm achieves a lower missed detection rate. At the 10\% missed detection rate threshold, our proposed algorithm provides a 3~dB SNR gain compared to the conventional algorithm. Fig.~7(c) illustrates our algorithm's superior false alarm suppression capability. Throughout the 6,000 simulation runs, the conventional algorithm generated 1,263 false alarms, while our proposed algorithm produced only 613 false alarms—a 51\% reduction. This improvement occurs because the conventional algorithm counts all detections exceeding the CFAR threshold as false alarms, whereas our proposed algorithm effectively filters out numerous noise-induced detections by leveraging the pattern of periodic peaks.

\subsection{Discussions}
The parameter values involved in our proposed algorithm depend on the 5G RS available during the sensing period. For example, when sensing with PRS as shown in Fig.~1(a), $N_\text{s}$ equals 3, while when using DMRS as shown in Fig.~1(b), $N_\text{s}$ equals 4. The selection of $N_\text{slot}$ depends on the velocity resolution and accuracy requirements, as well as the duration of the RSs. 

Compared to the conventional algorithm that requires only a single FFT, our proposed algorithm necessitates two FFT calculations at different frequencies, plus additional computational complexity of $\mathcal{O}((k_1+k_2)N_\text{s})$ to reduce missed detections and false alarms, where $k_1$ and $k_2$ represent the number of detections in the velocity profiles obtained at the two frequencies, respectively. This increase in computational complexity is acceptable considering the significant improvements in sensing performance.

In robot-aided ISAC, the estimated velocities of surrounding objects can support downstream robotic functions such as collision avoidance, target tracking, and cooperative perception. The proposed method is particularly suitable for connected robots because it can operate with irregular RS reuse opportunities induced by traffic-driven scheduling, without requiring any robot-specific waveform design or sensing-dedicated RSs.

\section{Conclusions}
This paper proposed a standard-compliant multi-periodogram method for velocity estimation in robot-aided ISAC by reusing irregular RSs. By showing that the resulting velocity profile decomposes into periodic peaks (set by the number of slots) and an amplitude-shaping term (set by the SP), we exploit the inherent multi-peak structure for robust detection. With multi-frequency peak differentiation and pattern-based validation, the proposed method mitigates ambiguities and suppresses noise detections in low SNR. Simulations show a 3~dB SNR gain at 10\% missed-detection rate and a 51\% reduction in false alarms versus conventional periodogram processing, without introducing dedicated SSs or modifying 3GPP standard.

\section{Acknowledgment}
This work is supported by Mobile Information Networks-National Science and Technology Major Project (No.2025ZD1302100) and the UK Engineering and Physical Sciences Research Council (EPSRC) grant EP/Y037243/1 for TITAN Telecoms Hub.

\bibliographystyle{IEEEtran}
\bibliography{ref} 

%\vspace{12pt}
\color{red}
\end{document}